\documentclass[10pt,conference,letterpaper]{IEEEtran}
\IEEEoverridecommandlockouts
\usepackage{cite}
\usepackage{amsmath,amssymb,amsfonts}
\usepackage{algorithmic}
\usepackage{graphicx}
\usepackage{textcomp}
\usepackage{xcolor}
\usepackage{multirow}
\usepackage{caption}
\usepackage{subcaption}
\usepackage{bbm}
\usepackage{float}
\usepackage{url}
\def\BibTeX{{\rm B\kern-.05em{\sc i\kern-.025em b}\kern-.08em
    T\kern-.1667em\lower.7ex\hbox{E}\kern-.125emX}}

\begin{document}

\bstctlcite{IEEEexample:BSTcontrol}

\title{Tangled Program Graphs as an alternative to DRL-based control algorithms for UAVs
\thanks{The work presented in this paper was supported by the AGH University of Krakow project no. 16.16.120.773. This research was funded, in whole or in part, by the Agence Nationale de la Recherche (ANR), grant ANR-22-CE25-0005-01. A CC BY license is applied to the AAM resulting from this submission, in accordance with  the open access conditions of the grant.}
}

\author{
\IEEEauthorblockN{Hubert Szolc}
\IEEEauthorblockA{\textit{Embedded Vision Systems Group}, \\ 
\textit{Department of Automatic Control} \\
\textit{and Robotics}, \\ 
\textit{AGH University of Krakow} \\
Kraków, Poland \\
hubert.szolc@agh.edu.pl}
\and
\IEEEauthorblockN{Karol Desnos}
\IEEEauthorblockA{\textit{Univ Rennes, INSA Rennes} \\
\textit{CNRS, IETR – UMR 6164}\\
F-35000 Rennes, France \\
karol.desnos@insa-rennes.fr}
\and
\IEEEauthorblockN{Tomasz Kryjak, Senior Member, IEEE}
\IEEEauthorblockA{\textit{Embedded Vision Systems Group}, \\ 
\textit{Department of Automatic Control} \\
\textit{and Robotics}, \\ 
\textit{AGH University of Krakow} \\
Kraków, Poland \\
tomasz.kryjak@agh.edu.pl}
}

\maketitle

\begin{abstract}
Deep reinforcement learning (DRL) is currently the most popular AI-based approach to autonomous vehicle control.
An agent, trained for this purpose in simulation, can interact with the real environment with a human-level performance.
Despite very good results in terms of selected metrics, this approach has some significant drawbacks: high computational requirements and low explainability.
Because of that, a DRL-based agent cannot be used in some control tasks, especially when safety is the key issue.
Therefore we propose to use Tangled Program Graphs (TPGs) as an alternative for deep reinforcement learning in control-related tasks.
In this approach, input signals are processed by simple programs that are combined in a graph structure.
As a result, TPGs are less computationally demanding and their actions can be explained based on the graph structure. 
In this paper, we present our studies on the use of TPGs as an alternative for DRL in control-related tasks.
In particular, we consider the problem of navigating an unmanned aerial vehicle (UAV) through the unknown environment based solely on the on-board LiDAR sensor.
The results of our work show promising prospects for the use of TPGs in control related-tasks.
\end{abstract}


\section{Introduction}

\begin{figure*}
    \centering
    \includegraphics[width=\textwidth]{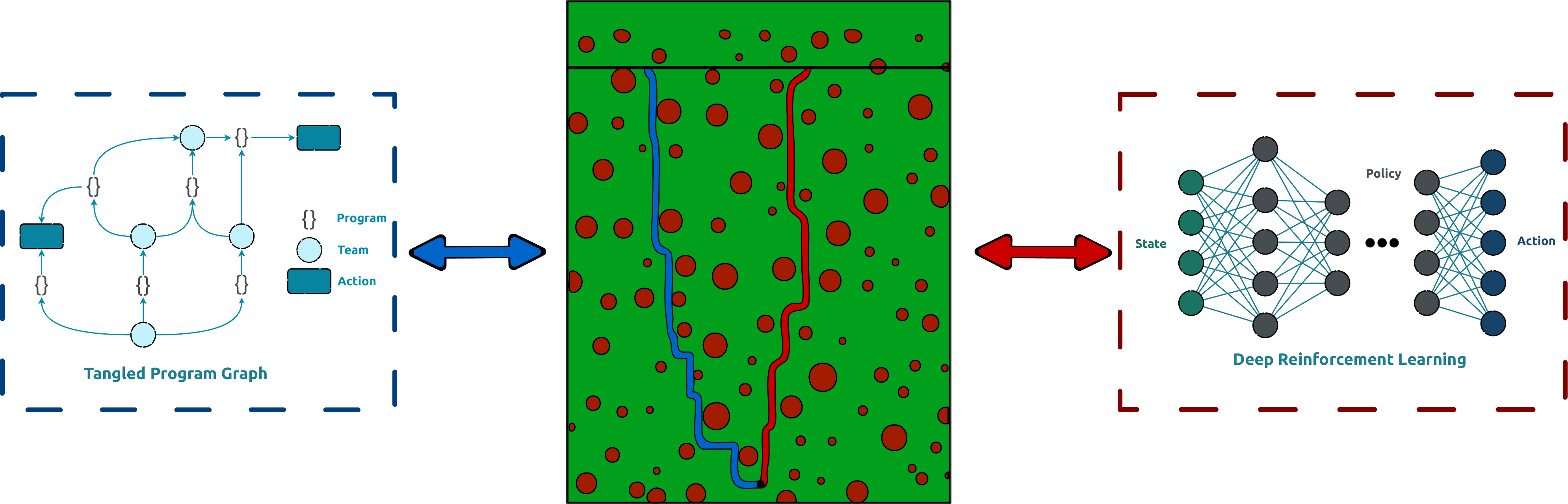}
    \caption{Tangled Program Graph (TPG) as an alternative for Deep Reinforcement Learning (DRL) in control related task of navigating the UAV through unknown forest environment.}
    \label{fig:teaser}
\end{figure*}

Deep reinforcement learning (DRL) has shown remarkable success in various applications; however, its implementation in autonomous control algorithms faces several significant limitations.
First of all, an agent trained this way often requires substantial computational resources to process data, which leads to challenges in real-time decision-making and responsiveness.
This is connected with memory constraints -- a typical issue of deep neural networks (DNNs), the core part of an agent -- and energy efficiency concerns.
Both of them are especially important for an unmanned aerial vehicle (UAV), where computing capabilities and power resources are strictly limited.
Another problem is the black-box nature of deep neural networks.
The lack of transparency in how DRL models make decisions can complicate the validation and verification processes, raising concerns about the trustworthiness of the autonomous control algorithms.

Taking into account aforementioned drawbacks of DRL, it seems reasonable to search for alternatives.
One promising possibility is the use of Tangled Program Graphs (TPGs)~\cite{Kelly2017}.
TPG is a directed graph acting as a control flow for the RL agent, defines relationships between environmental data and actions of the agent. 
This allows TPG-based models to effectively capture the nuances of state and action spaces without requiring large-scale data or intensive computational resources. 
They also facilitate more interpretable and efficient learning process.
At the same time, the training of TPG model can be implemented within the reinforcement learning framework.
This allows them to be directly compared in exactly the same environment.

In this paper we present the comparison between deep reinforcement learning and tangled program graphs in the task of navigating a UAV in a previously unknown environment (Fig.~\ref{fig:teaser}).
We have previously explored this problem with Proximal Policy Optimisation (PPO), an example of the deep reinforcement learning algorithm \cite{Miera2023}.
Therefore, we use a proprietary forest simulator, written in C++, to model these specific conditions.
It can be fully configurable with static (``normal'' trees) and/or moving (``fairytale'' trees) obstacles.
The first can be considered easier, the second more difficult.
The UAV is equipped with an on-board LiDAR sensor as the sole source of data for perceiving the environment.
As this is the first approach of its kind, we consider a somewhat simplified scenario.
Firstly, instead of the entire 3D space, we restrict the movement of the UAV to the XY plane (which corresponds to flight at a constant altitude).
Secondly, the dynamics of the drone are modelled as a point particle with limited acceleration and speed.
Thirdly, the obstacles (trees) are modelled using regular geometric shapes.
Despite the above limitations, the prepared environment remains a challenge for the autonomous control algorithm.
This allows quantitative evaluation and comparison of DRLs and TPGs in this task.

The main contribution of this paper can be summarised as follows:
\begin{itemize}
    \item To the best of our knowledge, this is the first attempt to use tangled program graphs to process LiDAR data and the first to use TPG to control an unmanned aerial vehicle.
    \item We demonstrate TPG as an alternative to deep reinforcement learning to obtain an agent capable of navigating a UAV through previously unknown environment.
    \item We present an in-depth comparison of tangled program graphs and deep reinforcement learning under different criteria.
    \item We develop a new environment for testing the DRL and TPG models.
    It is fully configurable and can be utilised to evaluate agents on maps with different levels of difficulty.
\end{itemize}
We make our entire code, including drone forest environment, available under an open licence in the GitHub repository\footnote{\url{https://github.com/vision-agh/drone_forest.git}}.

The remainder of this paper is organised as follows.
In Section~\ref{sec:related work} we describe DRL-based approaches for the autonomous control of unmanned aerial vehicles.
We also discuss previous attempts to compare tangled program graphs with reinforcement learning.
In Section~\ref{sec:tpg} we describe briefly the tangled program graphs.
In Section~\ref{sec:experiment setup}, we provide more detailed look at the used experimental setup.
In Section~\ref{sec:results} we present the obtained results along with a discussion.
The article concludes with a summary, in which we also outline plans for further work.

\section{Related work}
\label{sec:related work}

So far, promising results in autonomous UAV control can be achieved by using deep reinforcement learning \cite{Kaufmann2023}. 
An extensive overview of these approaches can be found in the work \cite{Jagannath2021} and -- in the context of autonomous drone racing -- in the article \cite{Hanover2024}. 
This is also the approach taken by the authors of the paper \cite{Rodriguez2019}, in which they consider the specific case of landing on a moving platform. 
For this task, a DRL agent equipped with two deep neural networks was utilised, for which the input was the state of the vehicle. 
To train them, the authors used the Deep Deterministic Policy Gradients (DDPG) algorithm running in the Gazebo simulator. 
However, after the training process, it was not decided to implement the algorithm on an embedded hardware platform. Instead, it was run on a ground station (in this case a laptop), while the calculated control was sent to the vehicle over a Wi-Fi connection. 
In this way, the trained algorithm was tested for different conditions (simulation and real-world) and its effectiveness was confirmed. 
Among the proposed future work, the authors point to the possibility of using different sources of the data for the input of the algorithm (instead of the current state of the vehicle).

Another example of using deep reinforcement learning for autonomous UAV control is presented in the paper \cite{Kaufmann2020}. 
It proposes an agent that enables extreme aerobatic manoeuvres that are challenging even for well-trained human pilots. 
Only images and data from the IMU were used as its input. 
After the training process, conducted in Gazebo simulator, the authors implemented the resulting control algorithm on an embedded Nvidia Jetson TX2 hardware platform. 
This gave the UAV a relatively high level of autonomy -- all data came from sensors mounted on the vehicle and was processed on-board.
This approach was further utilised by the authors for the drone racing, in which the trained agent achieved the level of human champions \cite{Kaufmann2023}, with results published in Nature journal.
As of today, this is probably the most impressive example of the use of DRL for the autonomous control of a UAV.

Despite the exceptionally good result in terms of control quality, the DRL suffers from a kind of black-box nature of the whole system.
The work \cite{Glanois2024} discusses different approaches to increase reinforcement learning interpretability.
The authors argue that this is in fact a multidimensional concept that can be associated with interpretable inputs, models and decisions.
Each of these appears to be central to the task of autonomous control.
However, despite numerous proposals, the lack of interpretability is still one of the key issues preventing DRL from becoming a more viable method for real-world applications.

As to date, to the best of our knowledge, there are no works regarding using TPG for the control of autonomous vehicle.
However, in the paper \cite{Kelly2017} they were utilised in the Atari 2600 video game environment.
The resulting TPG solutions demonstrated their high level of generalisation by achieving accuracy of deep learning models without requiring any more training resource than necessary for a single game title.
In the work \cite{Desnos2021} the new \texttt{GEGELATI} C++ library for TPGs was introduced.
As one of the demonstration examples, it was utilised to successfully train the agent in the simulation of classic control task -- stabilisation of inverted pendulum.

The related work presents deep reinforcement learning as a promising approach to achieve an autonomous control system for UAVs.
However, this comes with some limitations, in particular the need for sufficiently high computing power with low energy consumption.
Another major drawback is the lack of interpretability, which prevents further development of DRL-based solutions in real-world applications.
Therefore, the search for alternatives seems to be fully justified.

\section{Tangled Program Graphs}
\label{sec:tpg}

\begin{figure}
    \centering
    \includegraphics[width=0.38\textwidth]{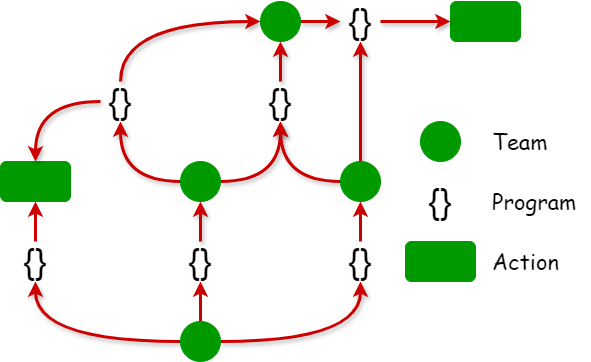}
    \caption{An example of tangled program graph (TPG).}
    \label{fig:tpg}
\end{figure}

The Tangled Program Graph (TPG) model, shown in Figure~\ref{fig:tpg}, is a directed graph composed of three key elements: programs, teams, and actions. 
In this structure, teams serve as internal nodes, and actions represent the leaves.
Programs are the edges of the graph, linking source teams to either destination teams or action vertices. 
Self-loops, or edges that connect a team to itself, are not permitted.
A program processes the current state of the environment to generate a real number, termed a bid. 
A program consists of a sequence of simple arithmetic instructions, such as additions or exponents. 
Each instruction uses operands from either the observed environment data or values stored in registers from previous instructions. 

The execution of a tangled program graph begins at its unique root team whenever a new environmental state is available. 
All programs connected to the root team are executed with the current state as their input. 
The execution then follows the edge with the highest bid, moving to the connected node. 
If this node is an another team, its outgoing programs are executed with the same state, and the process continues along the edge with the highest bid. 
This sequence repeats until the edge leads to an action vertex. 
At this point, the specified action is executed by the agent, the environment updates to a new state, and the TPG execution restarts from the root team.

The genetic evolution of a tangled program graph involves multiple root teams. 
Initially, the TPG for the first generation consists only of root teams with edges leading directly to action vertices. 
Each root team represents a distinct policy, whose fitness is evaluated by executing the TPG a set number of times or until a terminal state is reached. 
The rewards gathered during these evaluations determine the fitness of each root team. 
The genetic evolution process then eliminates the root teams with the lowest rewards from the TPG.

To generate new root teams for the next generation in the TPG evolution process, the remaining teams are randomly selected and duplicated along with their outgoing edges. 
These edges then undergo random mutations, which may alter their destination vertices and modify their programs by adding, removing, swapping, or changing instructions and operands. 
During mutation, surviving root teams from previous generations can become internal vertices if a new edge points to them. 
This process promotes the formation of valuable subgraphs that persist across generations, adding complexity to the TPG adaptively and enhancing the agent's rewards.

Compared to DRL, which uses a fixed decision--making topology with millions of parameters (deep networks), TPG is able to discover emergent representations suitable for each task.
This significantly reduces the computational complexity and memory requirements.
In addition, the graph-based representation, which encapsulates the dependencies and relationships within the environment, makes it possible to trace the steps that led to the execution of a specific action.
As a result, TPG is more explainable than DRL approaches.
A more detailed description of the tangled program graphs can be found in \cite{Kelly2018}.

\section{Experiment setup}
\label{sec:experiment setup}

For comparing tangled program graphs with deep reinforcement learning we utilise a proprietary simulator, written in C++.
It models the UAV equipped with a LiDAR sensor in the simplified forest environment.
This choice is motivated mainly by the following two factors:
\begin{enumerate}
    \item This is a kind of preliminary studies on using TPGs for control of an unmanned aerial vehicle.
    Therefore, we need a relatively simple environment, which can verify whether this approach is promising.
    \item In one of our previous works \cite{Miera2023}, we developed the first version of the simulator to test the LiDAR-based PPO agent.
    As a result, we had an initial version that only required integration with the TPG library.
\end{enumerate}
We describe the details of our simulator in the following Subsection~\ref{subsec:environment}.
It also takes into account significant differences between TPG and DRL training process.
We look at this in more depth in Subsection~\ref{subsec:tpg drl diff}.

\subsection{Drone forest environment}
\label{subsec:environment}

The drone forest environment, which we use in this work, is developed from the one described in our previous paper \cite{Miera2023}.
The UAV is equipped with an on-board 2D LiDAR sensor as the sole source of data for perceiving the environment.
The objective is to navigate the drone along the Y-axis for a predefined distance, avoiding collisions with obstacles.

\begin{figure}
    \centering
     \begin{subfigure}[b]{0.23\textwidth}
         \centering
         \includegraphics[width=\textwidth]{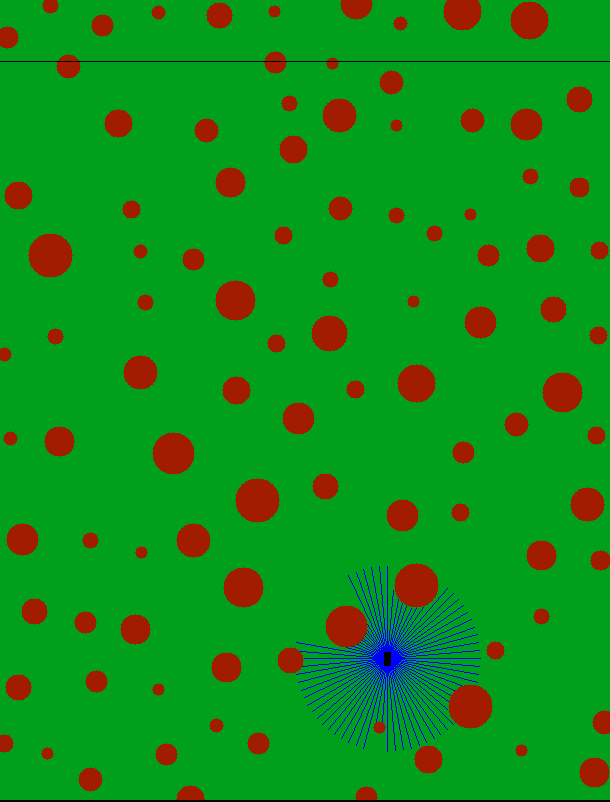}
     \end{subfigure}
     \hfill
     \begin{subfigure}[b]{0.23\textwidth}
         \centering
         \includegraphics[width=\textwidth]{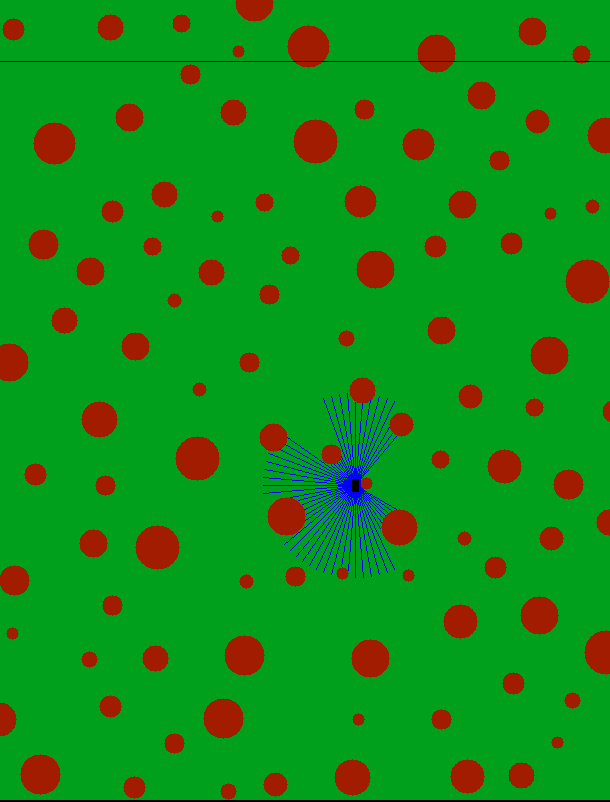}
     \end{subfigure}
    \caption{Two static configurations of the drone forest environment for two different seeds.}
    \label{fig:env static}
\end{figure}

The simulator models the vehicle dynamics as a point particle with limited acceleration and speed.
The drone can move only on the XY plane, which corresponds to flight at a constant altitude.
Furthermore, all objects are simulated as a simple geometric shapes: the UAV as a rectangle, the LiDAR as a bunch of rays, and the obstacles (``trees'') as circles. 
The environment is fully configurable, including forest size, drone dynamics limitations, LiDAR range, minimum spare distance between trees etc.
The placement and sizes of the obstacles are randomly selected based on the seed value.
An example of two random configurations of the environment is shown in Figure~\ref{fig:env static}.

The only observations available to the UAV control algorithm are the LiDAR outputs (distances measured by each beam).
Based on this, the agent can perform one of 40 actions.
They are defined as 10 values of the desired speed in one of the 4 directions (forward, left, backward, right).
It should be noted that due to its limited acceleration and inertia, a UAV may need some time to reach the speed of the given action.

\begin{figure}
    \centering
     \begin{subfigure}[b]{0.23\textwidth}
         \centering
         \includegraphics[width=\textwidth]{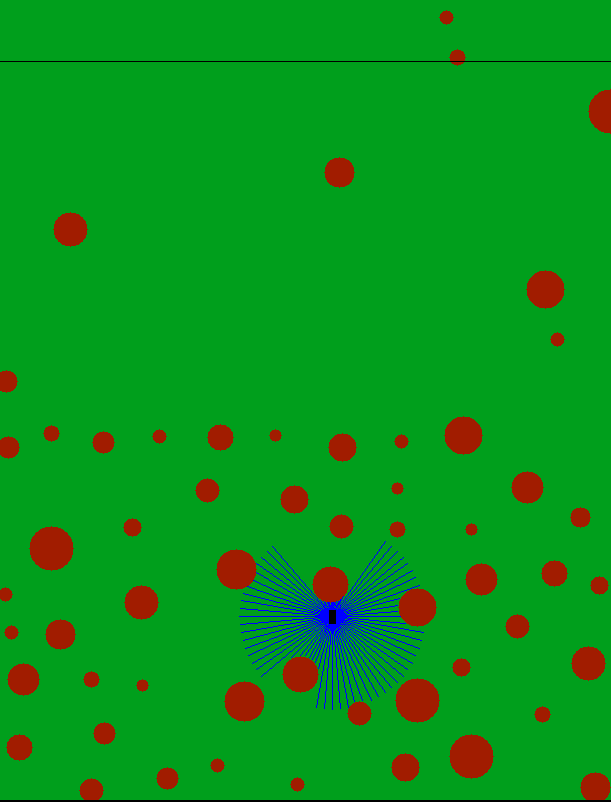}
     \end{subfigure}
     \hfill
     \begin{subfigure}[b]{0.23\textwidth}
         \centering
         \includegraphics[width=\textwidth]{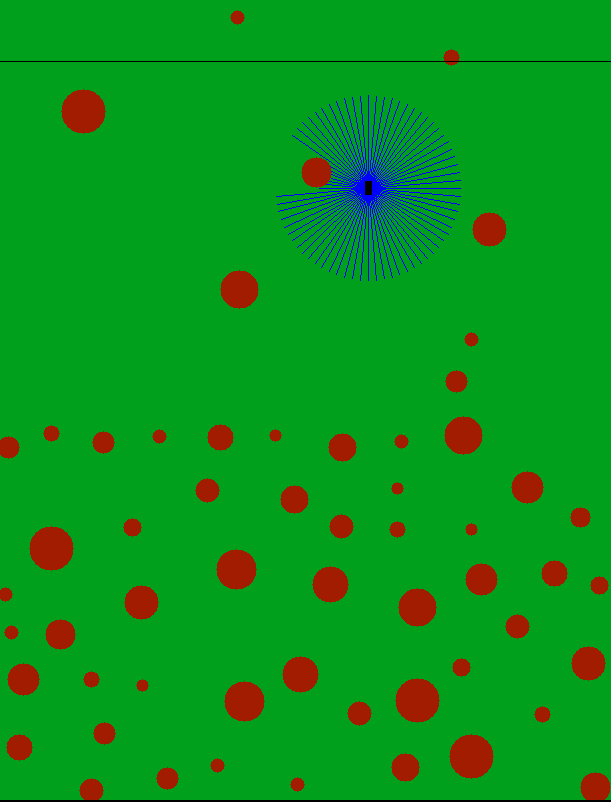}
     \end{subfigure}
    \caption{Dynamic configuration of the environment in two separate moments. Trees in the bottom part are static, while trees in the top part move horizontally with different speeds.}
    \label{fig:env dynamic}
\end{figure}

To provide more challenging conditions, the obstacles in the drone forest environment can be dynamic, i.e. they move horizontally between boundaries at a random velocity, which value is constant throughout the episode.
In this way it is possible to create ``hybrid'' maps (which we call the dynamic ones), with both static and moving obstacles.
An example of such a case is presented in Figure~\ref{fig:env dynamic}, in which trees in the bottom part are static, while trees in the top part move horizontally.

In order to enable seamless comparison between TPG and DRL agents, the drone forest environment includes two wrappers.
The first one connects it with the \texttt{GEGELATI}, which is the C++ library for training and evaluating TPG models \cite{Desnos2021}.
The second one binds the environment with \texttt{Stable-Baselines3}, which is a set of reliable implementations of reinforcement learning algorithms written in PyTorch \cite{stable-baselines3}.
In order to achieve data exchange between C++ and Python code, we use \texttt{pybind11} library \cite{pybind11}.

\subsection{TPG and DRL differences}
\label{subsec:tpg drl diff}

The aforementioned software wrappers allow the training process to be managed differently, depending on the requirements of the specific algorithm.
In DRL, the reward signal is used to adjust the weights of a neural network through back-propagation and gradient descent.
In particular, the PPO algorithm directly optimises the agent's policy based on the rewards collected.
TPG is an example of a genetic programming algorithm in which root teams (different agent policies) are evaluated based on their fitness.
Only the best survive the current generation and can then be linked together to form new policies.

Due to these differences, it is not guaranteed that the same reward function will give the best results for both DRL and TPG training.
Therefore, we decided to use the following reward function:
\begin{equation}
\label{eq:reward train}
    r_{training} = \begin{cases}
    -25 &\text{if} \quad \exists_i \quad U \cap T_i \neq \varnothing\\
    100 &\text{if} \quad y_{uav} \geq y_{goal}\\
    v_{alg} &\text{otherwise}
    \end{cases}
\end{equation}
where:
\begin{itemize}
    \item $U$ -- rectangle representing the UAV;
    \item $T_i$ -- circle representing the $i$-th tree;
    \item $y_{uav}$ -- Y coordinate of the current UAV position;
    \item $y_{goal}$ -- Y coordinate of the goal line;
    \item $v_{alg}$ -- value according to the algorithm, equals $v_{tpg}$ for the TPG or $v_{ppo}$ for the PPO.
\end{itemize}
Penalty for collision ($-25$) and reward for positive end of episode ($100$) were chosen by manual tuning.
For the TPG, $v_{alg}=v_{tpg}$ is simply the difference along Y axis between current UAV position and a goal:
\begin{equation}
\label{eq:valg tpg}
    v_{tpg} = y_{uav} - y_{goal}
\end{equation}
Note that the UAV always starts at the origin and the goal line is defined as $y=g \quad (g > 0)$.
Therefore, due to the form of \eqref{eq:valg tpg}, the $v_{alg}$ for TPG is a negative value that approaches 0 from the left as the UAV moves towards the goal.
On the other hand, for the PPO, $v_{alg}=v_{ppo}$ reflects the progress of UAV along Y axis between two consecutive steps:
\begin{equation}
\label{eq:valg ppo}
    v_{ppo} = \begin{cases}
        y_{uav}^k - y_{uav}^{k-1} &\text{if} \quad y_{uav}^k \geq y_{uav}^{k-1} \\
        2(y_{uav}^k - y_{uav}^{k-1}) &\text{otherwise}
    \end{cases}
\end{equation}
where: $y_{uav}^k$ -- UAV position after the step $k$.
Therefore, according to \eqref{eq:valg ppo}, the PPO agent is additionally penalised (by a factor of 2) for moving in the wrong direction.

During evaluation, we use the same reward for both agents:
\begin{equation}
\label{eq:reward eval}
    r_{evaluation} = y_{uav}
\end{equation}
In this way we measure the drone's progress towards the goal, which indicates the quality of the obtained agent (average distance covered during one episode).
We also count how many times the UAV reaches the goal line.
This value, in relation to all evaluation runs, is the accuracy of the agent.

\section{Results}
\label{sec:results}

We compare tangled program graphs with the proximal policy optimisation (PPO) algorithm as an example of deep reinforcement learning.
Firstly, we consider two configurations of a static environment: easier (with only 50 trees) and more difficult (with 100 trees).
Under these conditions we train both TPG and PPO agents and compare the best ones.
After that we also measure their transferability, i.e. in the environment with 50 trees we evaluate agents trained in environment with 100 trees and \textit{vice versa}.
Secondly, we train both agents in the more challenging dynamic environment, in which trees are static only in the first half of the forest.
Other obstacles move horizontally with random speeds.
In this configuration we test different sets of hyperparameters for the TPG training and compare the best one with the PPO agent.
In each case TPG agent and PPO agent are trained for 250 generations and 100 million steps, respectively.

\subsection{Static environment}

\begin{table}[]
    \centering
    \caption{Comparison between PPO and TPG trained and evaluated on static environment with the same number of trees.}
    \label{tab:static comparison}
    \begin{tabular}{|l|c||c|c|}
        \hline
        \textbf{Algorithm} & \textbf{Trees} & \textbf{Avg. distance [m]} & \textbf{Accuracy [\%]} \\
        \hline
        PPO & 50 & 21.10 & 93.00 \\
        \hline
        TPG & 50 & 20.74 & 90.00 \\
        \hline
        PPO & 100 & 19.55 & 83.00 \\
        \hline
        TPG & 100 & 20.71 & 92.00 \\
        \hline
    \end{tabular}
\end{table}

The evaluation results of the PPO and TPG agents in the static environment are shown in Table~\ref{tab:static comparison}.
TPG agent performs as well as its PPO rival in both cases.
In the easier configuration of 50 trees, it achieved an average distance of 20.74 [m] vs 21.10 [m] and an accuracy of 90\% vs 93\%.
In the more difficult environment, the TPG agent actually outperformed its rival, with an average distance of 20.71 [m] vs 19.55 [m] and an accuracy of 92\% vs 83\%.

\begin{table}[]
    \centering
    \caption{Comparison between PPO and TPG trained and evaluated on static environment with different number of trees.}
    \label{tab:static transfer}
    \begin{tabular}{|l|c||c|c|}
        \hline
        \textbf{Algorithm} & \textbf{Trees} & \textbf{Avg. distance [m]} & \textbf{Accuracy [\%]} \\
        \hline
        PPO & 50 $\rightarrow$ 100 & 19.28 & 72.00 \\
        \hline
        PPO & 100 $\rightarrow$ 50 & 21.08 & 93.00 \\
        \hline
        TPG & 50 $\rightarrow$ 100 & 16.71 & 53.00 \\
        \hline
        TPG & 100 $\rightarrow$ 50 & 21.57 & 97.00 \\
        \hline
    \end{tabular}
\end{table}

The results of the transferability test are gathered in Table~\ref{tab:static transfer}.
In the first case, when a model trained in a more difficult environment is evaluated in an easier one (i.e. transferring from 100 to 50 trees), both approaches perform well.
The PPO agent provides almost the same results as in the previous experiment (see Table~\ref{tab:static comparison}), while the TPG agent performs even better.
This is probably due to the randomness of the training, as the difference is not significant.
In the second case (i.e. transfer from 50 to 100 trees), the PPO agent outperforms its rival.
This is caused by the observed training trend of the TPG model.
When avoiding the obstacle, it prefers one horizontal direction over another.
In the easier environment, with less number of trees, it does not affect the final outcome.
However, with more obstacles on the map, TPG agent performs higher number of avoiding manoeuvres and flies outside the horizontal bounds.

\subsection{Dynamic environment}

\begin{table*}[]
    \centering
    \caption{Comparison of different TPG training hyperparameters in the dynamic environment. The names of the hyperparameters are derived from the \texttt{GEGELATI} library -- their meaning and details can be found in the documentation of that library.}
    \label{tab:dynamic tpg}
    \begin{tabular}{|c|c|c|c|c|c|c|c||c|c|c|c|c|c|c|c||c|c|}
        \hline
        \multicolumn{8}{|c||}{Program mutation} & \multicolumn{8}{c||}{Graph mutation} & & \\
        \hline
         \rotatebox{90}{maxConstValue} & \rotatebox{90}{maxProgramSize} & \rotatebox{90}{minConstValue} & \rotatebox{90}{pAdd} & \rotatebox{90}{pConstantMutation} & \rotatebox{90}{pDelete} & \rotatebox{90}{pMutate} & \rotatebox{90}{pSwap} & \rotatebox{90}{maxInitOutEdges} & \rotatebox{90}{maxOutEdges} & \rotatebox{90}{nbRoots} & \rotatebox{90}{pEdgeAddition} & \rotatebox{90}{pEdgeDeletion} & \rotatebox{90}{pEdgeDestChange} & \rotatebox{90}{pEdgeDestIsAction} & \rotatebox{90}{pProgramMutation} & \rotatebox{90}{Avg. distance [m]} & \rotatebox{90}{Accuracy [\%]}\\
        \hline
        50 & 96 & -20 & 0.5 & 0.5 & 0.5 & 0.7 & 0.7 & 3 & 5 & 360 & 0.7 & 0.7 & 0.1 & 0.5 & 0.2 & 14.72 & 43.00\\
        \hline
        50 & 96 & -20 & 0.5 & 0.5 & 0.5 & 0.7 & 0.7 & 3 & 5 & 216 & 0.7 & 0.7 & 0.1 & 0.5 & 0.2 & 18.00 & 58.00\\
        \hline
        50 & 96 & -20 & 0.5 & 0.5 & 0.5 & 0.7 & 0.7 & 3 & 5 & 252 & 0.7 & 0.7 & 0.1 & 0.5 & 0.2 & 17.65 & 55.00\\
        \hline
        \textbf{50} & \textbf{96} & \textbf{-20} & \textbf{0.5} & \textbf{0.5} & \textbf{0.5} & \textbf{0.7} & \textbf{0.7} & \textbf{3} & \textbf{5} & \textbf{288} & \textbf{0.7} & \textbf{0.7} & \textbf{0.1} & \textbf{0.5} & \textbf{0.2} & \textbf{18.76} & \textbf{65.00}\\
        \hline
        100 & 96 & -10 & 0.5 & 0.5 & 0.5 & 1.0 & 1.0 & 3 & 5 & 288 & 0.7 & 0.7 & 0.1 & 0.5 & 0.2 & 16.30 & 48.00\\
        \hline
        100 & 96 & -10 & 0.5 & 0.5 & 0.5 & 0.7 & 1.0 & 3 & 5 & 288 & 0.7 & 0.7 & 0.1 & 0.5 & 0.2 & 15.32 & 47.00\\
        \hline
        100 & 96 & -10 & 0.5 & 0.5 & 0.5 & 0.7 & 0.7 & 3 & 5 & 288 & 0.7 & 0.7 & 0.1 & 0.5 & 0.2 & 16.82 & 51.00\\
        \hline
        100 & 96 & -10 & 0.5 & 0.5 & 0.5 & 0.7 & 0.7 & 3 & 5 & 288 & 0.8 & 0.7 & 0.1 & 0.5 & 0.2 & 16.48 & 55.00\\
        \hline
        100 & 96 & -10 & 0.5 & 0.5 & 0.5 & 0.7 & 0.7 & 3 & 5 & 288 & 0.8 & 0.8 & 0.1 & 0.5 & 0.2 & 12.89 & 30.00\\
        \hline
        100 & 96 & -10 & 0.5 & 0.5 & 0.5 & 0.7 & 0.7 & 3 & 5 & 288 & 0.8 & 0.6 & 0.1 & 0.5 & 0.2 & 13.44 & 40.00\\
        \hline
        100 & 96 & -10 & 0.5 & 0.5 & 0.5 & 0.7 & 0.7 & 3 & 5 & 288 & 0.8 & 0.7 & 0.1 & 0.3 & 0.2 & 8.64 & 12.00\\
        \hline
        100 & 96 & -10 & 0.5 & 0.5 & 0.5 & 0.7 & 0.7 & 3 & 5 & 324 & 0.8 & 0.7 & 0.1 & 0.5 & 0.2 & 17.11 & 56.00\\
        \hline
    \end{tabular}
\end{table*}

In the dynamic environment we firstly test different sets of hyperparameters for the TPG training.
The results are shown in Table~\ref{tab:dynamic tpg}.
Noticeably, small changes in some parameters (e.g. \texttt{pEdgeDestIsAction}, \texttt{pEdgeDeletion}, or \texttt{pEdgeAddition}) can lead to significant changes in the agent performance.
In general, it seems that graph mutation parameters are the determining factors for the training process, while program mutation parameters can be used for a kind of ``fine tuning''.

\begin{table}[]
    \centering
    \caption{Comparison between PPO and TPG trained and evaluated on the dynamic environment.}
    \label{tab:dynamic comparison}
    \begin{tabular}{|l||c|c|}
        \hline
        \textbf{Algorithm} & \textbf{Avg. distance [m]} & \textbf{Accuracy [\%]} \\
        \hline
        PPO & 20.42 & 83.00 \\
        \hline
        TPG & 18.76 & 65.00 \\
        \hline
    \end{tabular}
\end{table}

In Table~\ref{tab:dynamic comparison} we present the comparison of the best model from the previous experiment (see Table~\ref{tab:dynamic tpg}) with the PPO agent.
The results presented are from the experiments performed on the exact same set of 100 scenarios.
In this case, somehow different from the static environment, the DRL-based agent performs much better than the TPG model.
We do not see an obvious reason for this result.
Therefore, this topic needs further research, in particular the in-depth analysis of the influence of the form of the reward function on the resulting agent. 

\section{Summary}
\label{sec:summary}

In this paper, we presented our studies on the use of tangled program graphs as an alternative to deep reinforcement learning in control-related tasks.
In particular, we considered the problem of navigating an unmanned aerial vehicle through the unknown environment based solely on the on-board LiDAR sensor.
To made the aforementioned comparison, we had developed a new drone forest environment.
It is written in the C++ language, can be configured depending on the needs and allows creating maps with different levels of difficulty.

We trained and evaluated TPG and PPO agents on both static and dynamic configurations of the environment.
In the first one, we considered two numbers of trees: 50 and 100.
In both cases, utilising the TPG and PPO models gives similar results.
We also verified the transferability of both models between aforementioned static cases.
In the dynamic environment we tested different sets of hyperparameters for TPG training.
Based on that, we selected the best model and compared it with PPO agent. 

The results of our work show promising prospects for the use of TPGs in control-related tasks.
The TPG agent matches the performance of the PPO rival in a static environment.
However, it performs worse in the more challenging scenario with moving obstacles.
As we do not see an obvious reason for such a result, there is still a lot of work to be done.
First of all, an in-depth analysis of the influence of the form of the reward function on the resulting agent should be prepared in order to achieve the best possible performance of the TPG agent.
Another problem is to find the best set of hyperparameters using some kind of optimisation algorithm.
However, this task should be much easier than in DRL due to the lower dimension of the hyperparameter space.

Other directions of future work include more complex environments prepared with the use of simulation environments (e.g. Gazebo, Nvidia Isaac Sim).
After positive verification in such simulators, the TPG agent could be deployed on the embedded computing platform, such as a SoC FPGA (system on a chip field-programmable gate array) device, and run on the UAV.
To achieve the best control quality, TPG could also process data from various sensors, such as frame-based or event cameras.

\bibliographystyle{IEEEtran.bst}
\bibliography{IEEEfull,bibliography.bib}

\end{document}